\documentclass[10pt,twocolumn,letterpaper]{article}
\usepackage{cvpr}
\usepackage{times}
\usepackage{epsfig}
\usepackage{graphicx}
\usepackage{amsmath}
\usepackage{amssymb}
\usepackage{color}
\usepackage{amsmath}
\usepackage{algorithm}
\usepackage{subfig}
\usepackage[noend]{algpseudocode}
\usepackage{color}
\usepackage{subfig}
\usepackage{multicol}
\usepackage{enumitem}
\usepackage[font={small}]{caption}
\DeclareGraphicsExtensions{.pdf,.png, .PNG, .jpg. JPG, .jpeg, .JPEG}
\usepackage[colorlinks]{hyperref}
\hypersetup{
    citecolor=green,
    linkcolor=red,   
    urlcolor=magenta}

\cvprfinalcopy 
\def\cvprPaperID{2441} % *** Enter the CVPR Paper ID here

\newcommand{\comm}[1]{}

\begin{document}

\title{
What makes ImageNet good for transfer learning? \\
} 

\comm{ % Uncomment for individual labeling
    \author{Minyoung Huh \\
    UC Berkeley\\
    {\tt\small minyoung@berkeley.edu}
    \and
    Pulkit Agrawal\\
    UC Berkeley\\
    {\tt\small pulkitag@berkeley.edu}
    \and
    Alexei A. Efros\\
    UC Berkeley\\
    {\tt\small aaefros@berkeley.edu}
    }
}

\author{Minyoung Huh \qquad Pulkit Agrawal \qquad Alexei A. Efros\\
Berkeley Artificial Intelligence Research (BAIR) Laboratory\\
UC Berkeley\\
{\tt\small \{minyoung,pulkitag,aaefros\}@berkeley.edu  }
}

\twocolumn[
   \begin{center}
    \maketitle
   \end{center}
]
\begin{abstract}
\comm{
%David's suggestion
The tremendous success of ImageNet-trained deep features on a wide range of transfer tasks begs the question: what are the properties of the ImageNet dataset that are critical for learning good, general-purpose features? This work provides an empirical investigation of various facets of this question: 
Is more pre-training data always better?
How does feature quality depend on the number of training examples per class?
Does adding more object classes improve performance?
For the same data budget, how should the data be split into classes?
Is fine-grained recognition necessary for learning good features?
Given the same number of training classes, is it better to have coarse classes or fine-grained classes?  Which is better: more classes or more examples per class? 
To answer these and related questions, we pre-trained CNN features on various subsets of the ImageNet dataset and evaluated transfer performance on PASCAL detection, PASCAL action classification, and SUN scene classification tasks. 
Our overall findings suggest that most changes in the choice of pre-training data long thought to be critical do not significantly affect transfer performance.
}
%The tremendous success of features learnt using the ImageNet classification task on a wide range of transfer tasks raises the question: 
The tremendous success of ImageNet-trained deep features on a wide range of transfer tasks raises the question: 
what is it about the ImageNet dataset that makes the learnt features as good as they are?
%what properties of the ImageNet dataset are critical for learning good, general-purpose features? 
This work provides an empirical investigation into the various facets of this question, such as, looking at the importance of the amount of examples, number of classes, balance between images-per-class and classes, and the role of fine and coarse grained recognition. We pre-train CNN features on various subsets of the ImageNet dataset and evaluate transfer performance on a variety of standard vision tasks. Our overall findings suggest that most changes in the choice of pre-training data long thought to be critical, do not significantly affect transfer performance.
\end{abstract}

\section{Introduction}
%\blfootnote{Our work can be found on our website \\ \tt{\url{http://minyounghuh.com/papers/analysis}}}

It has become increasingly common within the computer vision community to treat image classification on ImageNet~\cite{ILSVRC15} not as an end in itself, but rather as a ``pretext task" for training deep convolutional neural networks (CNNs~\cite{lecun1989backpropagation,krizhevsky2012ImageNet}) to learn good general-purpose features. This practice of first training a CNN to perform image classification on ImageNet (i.e. pre-training) and then adapting these features for a new target task (i.e. fine-tuning) has become the de facto standard for solving a wide range of computer vision problems. Using ImageNet pre-trained CNN features, impressive results have been obtained on several image classification datasets~\cite{donahue2013decaf,razavian2014cnn}, as well as object detection~\cite{girshick2014rich,sermanet2013overfeat}, action recognition~\cite{simonyan2014two}, human pose estimation~\cite{carreira2015human}, image segmentation~\cite{dai2015instance}, optical flow~\cite{weinzaepfel2013deepflow}, image captioning~\cite{donahue2015long,karpathy2015deep} and others ~\cite{lecun2015deep}.

Given the success of ImageNet pre-trained CNN features, it is only natural to ask: what is it about the ImageNet dataset that makes the learnt features as good as they are?  One school of thought believes that it is the sheer size of the dataset (1.2 million labeled images) that forces the representation to be general.  Others argue that it is the large number of distinct object classes (1000), which forces the network to learn a hierarchy of generalizable features. Yet others believe that the secret sauce is not just the large number of classes, but the fact that many of these classes are visually similar (e.g. many different breeds of dogs), turning this into a fine-grained recognition task and pushing the representation to ``work harder''. But, while almost everyone in computer vision seems to have their own opinion on this hot topic, little empirical evidence has been produced so far.

In this work, we systematically investigate which aspects of the ImageNet task are most critical for learning good general-purpose features. We evaluate the features by fine-tuning on three tasks: object detection on PASCAL-VOC 2007 dataset (PASCAL-DET), action classification on PASCAL-VOC 2012 dataset (PASCAL-ACT-CLS) and scene classification on the SUN dataset (SUN-CLS); see Section \ref{sec:setup} for more details. 

%NIPS STYLE FIGURE 1 + 2
\begin{figure*}
\centering
\begin{minipage}{.49\textwidth}
  \centering
  \includegraphics[width=1.0\linewidth]{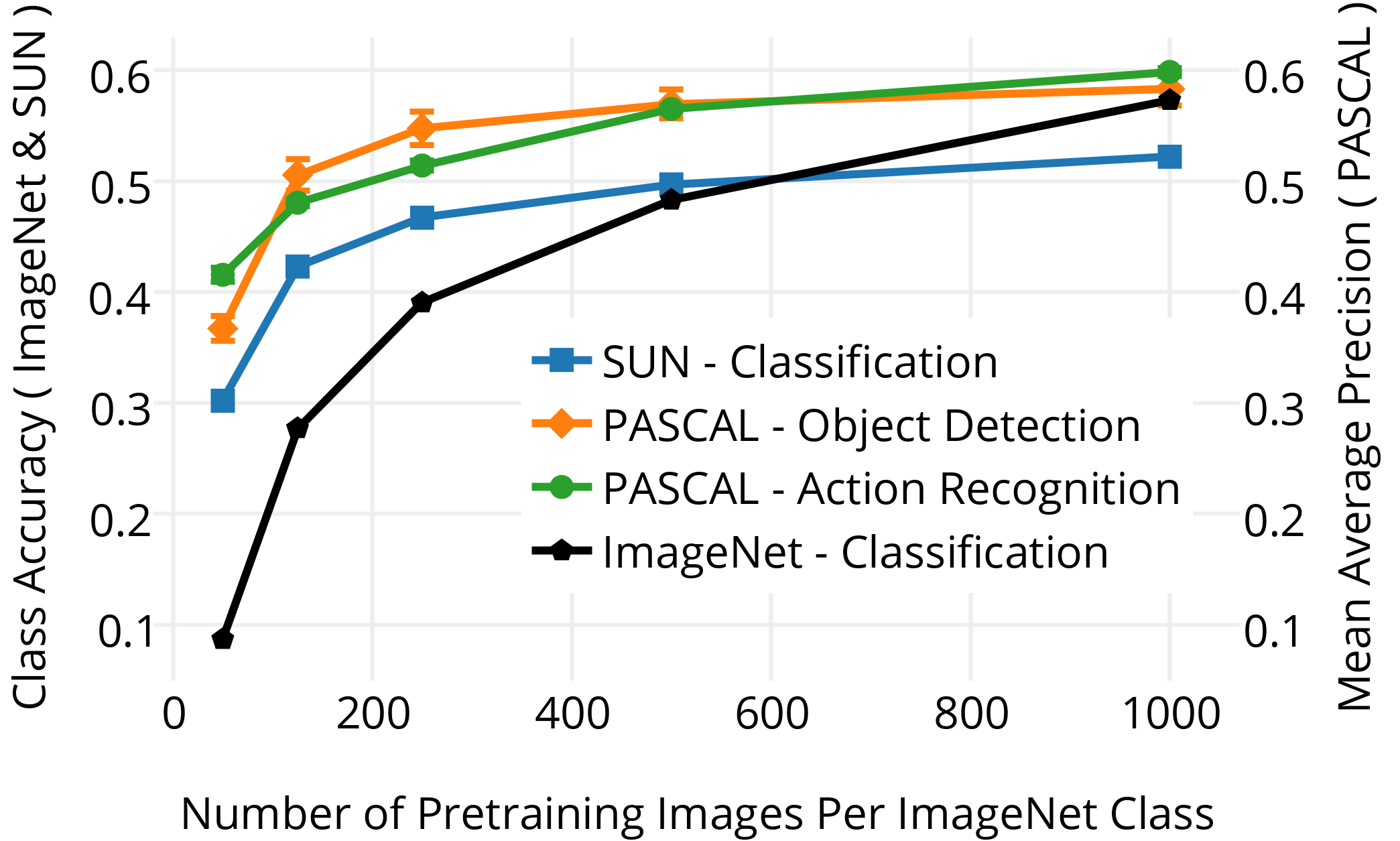}
  \captionof{figure}{Change in transfer task performance of a CNN pre-trained with varying number of images per ImageNet class. The left y-axis is the mean class accuracy used for SUN and ImageNet CLS. The right y-axis measures mAP for PASCAL DET and ACTION-CLS. The number of examples per class are reduced by random sampling. Accuracy on the ImageNet classification task increases faster as compared to performance on transfer tasks.}
  \label{fig:number-of-class}
\end{minipage}%
\hspace{5px}
\begin{minipage}{.49\textwidth}
  \centering
  \includegraphics[width=1.0\linewidth]{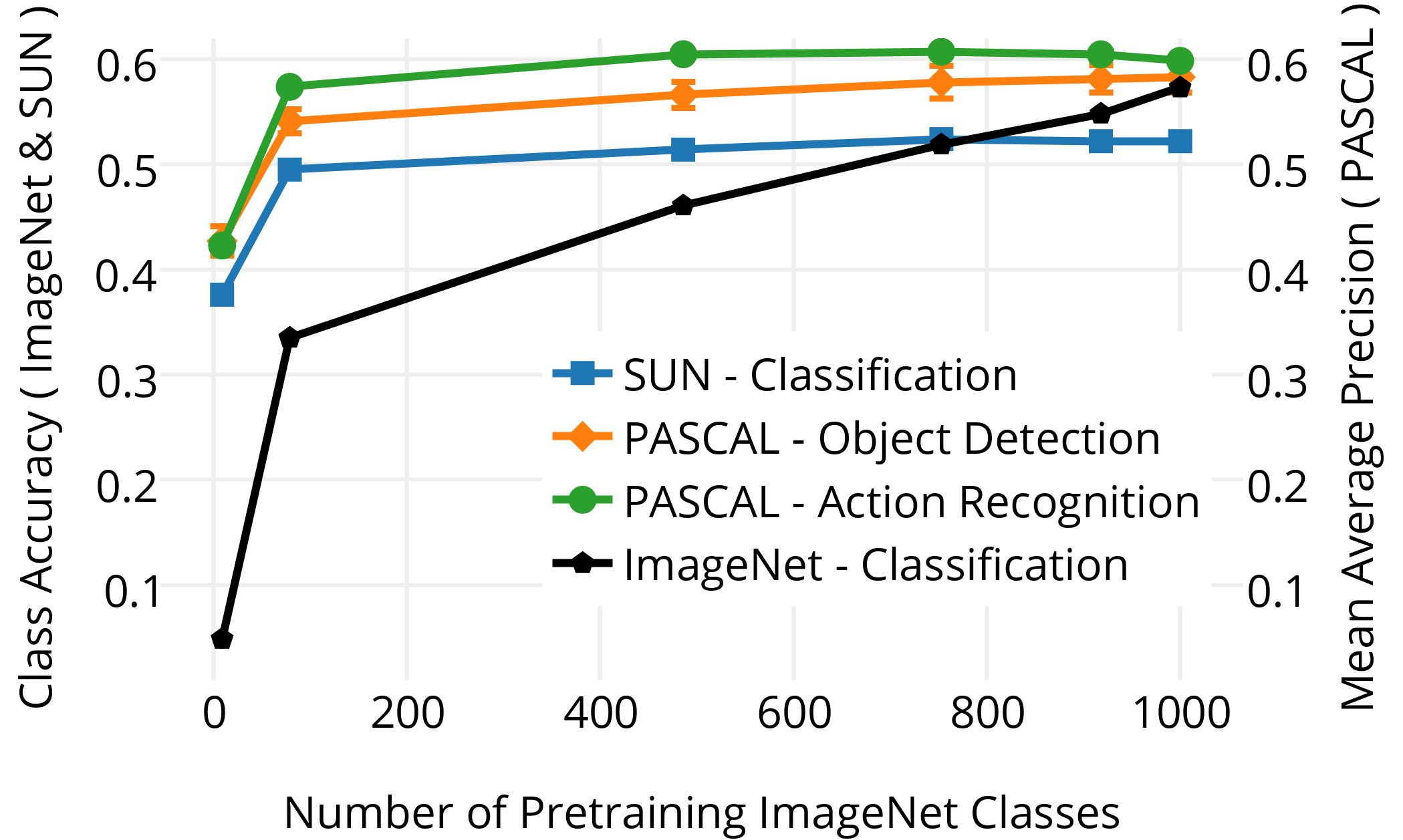}
  \captionof{figure}{Change in transfer task performance with varying number of pre-training ImageNet classes. The number of ImageNet classes are varied using the technique described in Section \ref{sec:finegrain}. With only 486 pre-training classes, transfer performances are unaffected and only a small drop is observed when only 79 classes are used for pre-training. The ImageNet classification performance is measured by fintetuning the last layer to the original 1000-way classification.}
  \label{fig:hierarchy}
\end{minipage}
%\vspace{-0.1in}
\end{figure*}

The paper is organized as a set of experiments answering a list of key questions about feature learning with ImageNet. The following is a summary of our main findings: 

\medskip

\noindent\textit{1. How many pre-training ImageNet examples are sufficient for transfer learning?} Pre-training with only \textit{half} the ImageNet data (500 images per class instead of 1000) results in only a small drop in transfer learning performance (1.5 mAP drop on PASCAL-DET). This drop is much smaller than the drop on the ImageNet classification task itself. 
See Section~\ref{sec:examples_per_class} and Figure~\ref{fig:number-of-class} for details. 

\medskip

\noindent \textit{2. How many pre-training ImageNet classes are sufficient for transfer learning?} 
Pre-training with an order of magnitude \textit{fewer} classes (127 classes instead of 1000) results in only a small drop in transfer learning performance (2.8 mAP drop on PASCAL-DET). Curiously, we also found that for some transfer tasks, pre-training with fewer classes leads to better performance. 
See Section~\ref{sec:finegrain} and Figure~\ref{fig:hierarchy} for details. 

\medskip

\noindent \textit{3. How important is fine-grained recognition for learning good features for transfer learning?} Features pre-trained with a subset of ImageNet classes that do not require fine-grained discrimination still demonstrate good transfer performance. See Section \ref{sec:fine-necessary} and Figure~\ref{fig:hierarchy} for details. 
\medskip

\noindent \textit{4. Does pre-training on coarse classes produce features capable of fine-grained recognition (and vice versa) on ImageNet itself?}  We found that a CNN trained to classify only between the 127 coarse ImageNet classes produces features capable of telling apart fine-grained ImageNet classes whose labels it has never seen in training (section~\ref{sec:relevant}). Likewise, a CNN trained to classify the 1000 ImageNet classes is able to distinguish between unseen coarse-level classes higher up in the WordNet hierarchy (section~\ref{sec:relevant2}).
\medskip

\noindent \textit{5. Given the same budget of pre-training images, should we have more classes or more images per class?} Training with fewer classes but more images per class performs slightly better at transfer tasks than training with more classes but fewer images per class. See Section~\ref{sec:mc_or_me} and Table~\ref{table:class-vs-example} for details. 
\medskip

\noindent \textit{6. Is more data always helpful?} We found that training with 771 ImageNet classes (out of 1000) that \textit{exclude} all PASCAL VOC classes, achieves nearly the same performance on PASCAL-DET as training on complete ImageNet. Further experiments confirm that blindly adding more training data does \textit{not} always lead to better performance and can sometimes hurt performance. See Section \ref{sec:more_class}, and Table \ref{figure:split} for more details. 

\section{Related Work}
A number of papers have studied transfer learning in CNNs, including the various factors that affect pre-training and fine-tuning.  For example, the question of whether pre-training should be terminated early to prevent over-fitting and what layers should be used for transfer learning was studied by~\cite{agrawal2014analyzing,yosinski2014transferable}. A thorough investigation of good architectural choices for transfer learning was conducted by~\cite{azizpour2015generic}, while \cite{li2016withoutforgetting} propose an approach to fine-tuning for new tasks without ''forgetting'' the old ones.  In contrast to these works, we use a fixed fine-tuning pr

One central downside of supervised pre-training is that large quantity of expensive manually-supervised training data is required. The possibility of using large amounts of unlabelled data for feature learning has therefore been very attractive. Numerous methods for learning features by optimizing some auxiliary criterion of the data itself have been proposed. The most well-known such criteria are image reconstruction ~\cite{bourlard1988auto,salakhutdinov2009deep,olshausen1996emergence,mobahi2009deep,ranzato2007unsupervised, kingma2013auto} (see \cite{bengio2012unsupervised} for a comprehensive overview) and feature slowness~\cite{wiskott2002slow,goroshin2015unsupervised}.  Unfortunately, features learned using these methods turned out not to be competitive with those obtained from supervised ImageNet pre-training~\cite{pathak2016context}. To try and force better feature generalization, more recent ``self-supervised'' methods use more difficult data prediction auxiliary tasks in an effort to make the CNNs ``work harder''.  Attempted self-supervised tasks include predictions of  ego-motion~\cite{agrawal2015learning,jayaraman2015learning}, spatial context~\cite{doersch2015unsupervised,pathak2016context,noroozi2016puzzles}, temporal context~\cite{wang2015unsupervised}, and even color~\cite{zhang2016color,larsson2016learning} and sound~\cite{owens2016sound}. While features learned using these methods often come close to ImageNet performance, to date, none have been able to beat it. 
    
A reasonable middle ground between the expensive, fully-supervised pre-training and free unsupervised pre-training is to use weak supervision.  For example, \cite{armand2015weak} use the YFCC100M dataset of 100 million Flickr images labeled with noisy user tags as pre-training instead of ImageNet.  But yet again, even though YFCC100M is almost two orders of magnitude larger than ImageNet, somewhat surprisingly, the resulting features do not appear to give any substantial boost over these pre-trained on ImageNet.
     
Overall, despite keen interest in this problem, alternative methods 
for learning general-purpose deep features have not managed to  outperform ImageNet-supervised pre-training on transfer tasks. 

The goal of this work is to try and understand what is the secret to ImageNet's continuing success.

\begin{figure}[t]
    \begin{center}
    \includegraphics[width=1.0\linewidth]{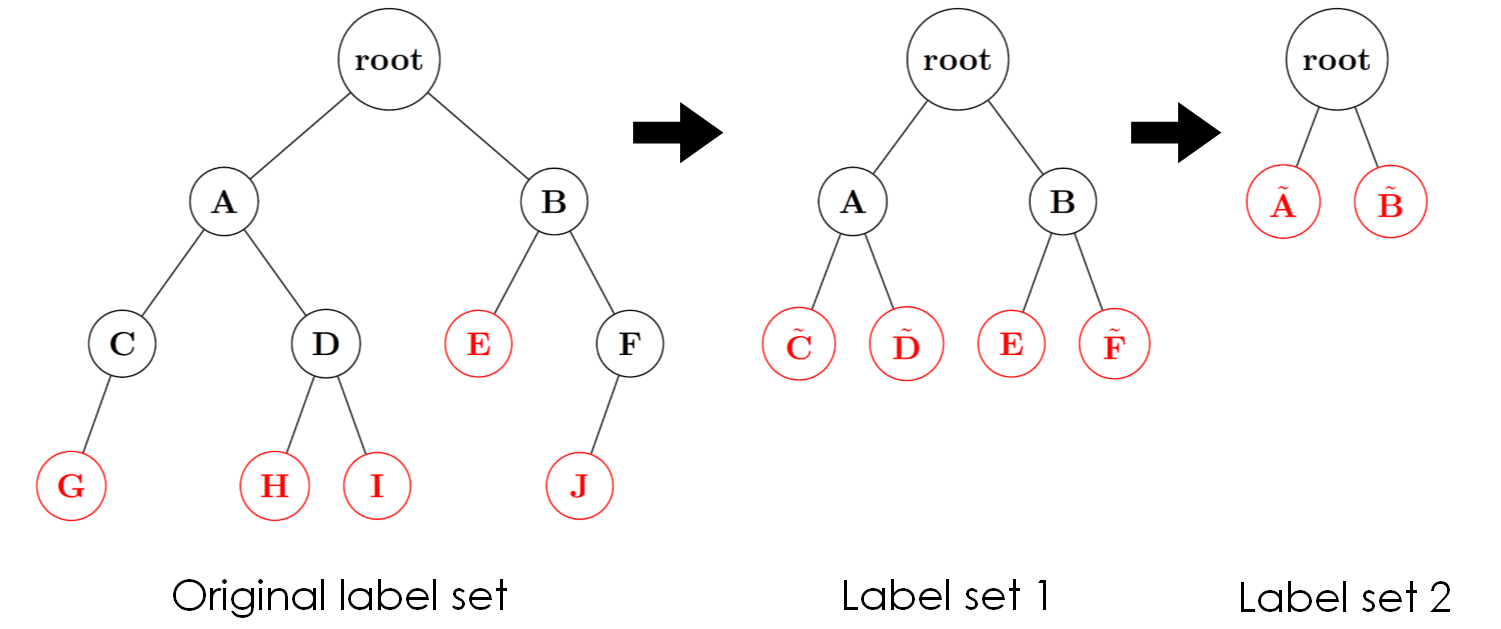}\\
    \vspace{-.1in}
    \caption{An illustration of the bottom up procedure used to construct different label sets using the WordNet tree. Each node of the tree represents a class and the leaf nodes are shown in red. Different label sets are iteratively constructed by clustering together all the leaf nodes with a common parent. In each iteration, only leaf nodes are clustered. This procedure results into a sequence of label sets for 1.2M images, where each consequent set contains labels coarser than the previous one. Because the WordNet tree is imbalanced, even after multiple iterations, label sets contain some classes that are present in the 1000 way ImageNet challenge.}
    \vspace{-.2in}
    \label{fig:imgnttree}
    \end{center}
\end{figure}

\section{Experimental Setup}
\label{sec:setup}
The process of using supervised learning to initialize CNN parameters using the task of ImageNet classification is referred to as pre-training. The process of adapting pre-trained CNN to continuously train on a target dataset is referred to as finetuning. All of our experiments use the Caffe~\cite{caffe} implementation of the a single network architecture proposed by Krizhevsky et al.~\cite{krizhevsky2012ImageNet}. We refer to this architecture as AlexNet.

We closely follow the experimental setup of Agrawal et al.~\cite{agrawal2014analyzing} for evaluating the generalization of pre-trained features on three transfer tasks: PASCAL VOC 2007 object detection (PASCAL-DET), PASCAL VOC 2012 action recognition (PASCAL-ACT-CLS) and scene classification on SUN dataset (SUN-CLS). 

\begin{itemize}[leftmargin=*]
\item For PASCAL-DET, we used the PASCAL VOC 2007 train/val for finetuning using the experimental setup and code provided by Faster-RCNN~\cite{ren2015faster} and report performance on the test set. Finetuning on PASCAL-DET was performed by adapting the pre-trained convolution layers of AlexNet. The model was trained for 70K iterations using stochastic gradient descent (SGD), with an initial learning rate of 0.001 with a reduction by a factor of 10 at 40K iteration.

\item For PASCAL-ACT-CLS, we used PASCAL VOC 2012 train/val for finetuning and testing using the experimental setup and code provided by R*CNN ~\cite{rstarcnn2015}. The finetuning process for PASCAL-ACT-CLS mimics the procedure described for PASCAL-DET.

\item For SUN-CLS we used the same train/val/test splits as used by ~\cite{agrawal2014analyzing}. Finetuning on SUN was performed by first replacing the FC-8 layer in the AlexNet model with a randomly initialized, and fully connected layer with 397 output units. Finetuning was performed for 50K iterations using SGD with an initial learning rate of 0.001 which was reduced by a factor of 10 every 20K iterations.
\end{itemize}

Faster-RCNN and R*CNN are known to have variance across training runs; we therefore run it three times and report the mean $\pm$ standard deviation. On the other hand, \cite{agrawal2014analyzing}, reports little variance between runs on SUN-CLS so we report our result using a single run.

In some experiments we pre-train on ImageNet using a different number of images per class. The model with 1000 images/class uses the original ImageNet ILSVRC 2012 training set. Models with N images/class for $N < 1000$ are trained by drawing a random sample of N images from all images of that class made available as part of the ImageNet training set.

\setlength{\tabcolsep}{4pt}
\begin{table}[t]
\begin{center}
\scalebox{0.97}{
    \begin{tabular}{l|ccc}
    \hline
    \; Pre-trained Dataset \; & \qquad PASCAL & \qquad SUN \qquad \qquad&\\
    \hline
    \; Original \; & \qquad 58.3 & \qquad 52.2 \qquad \qquad&\\
    \; 127 Classes \;& \qquad 55.5 & \qquad  48.7 \qquad \qquad&\\
    \hline
    \; Random \; & \qquad 41.3 \cite{philipp2016init} & \qquad 35.7 ~\cite{agrawal2014analyzing} \qquad \qquad &\\
    \hline
    \end{tabular}
}
%\vspace{5px}
\vspace{-.1in}
\caption{The transfer performance of a network pre-trained using 127 (coarse) classes obtained after top-down clustering of the WordNet tree is comparable to a transfer performance after finetuning on all 1000 ImageNet classes. This indicates that fine-grained recognition is not necessary for learning good transferable features.}
\vspace{-.2in}
\label{table:coarse}
\end{center}
\end{table}
\setlength{\tabcolsep}{1.4pt}

\begin{figure*}[t!]
\begin{center}
\includegraphics[width=0.495\linewidth]{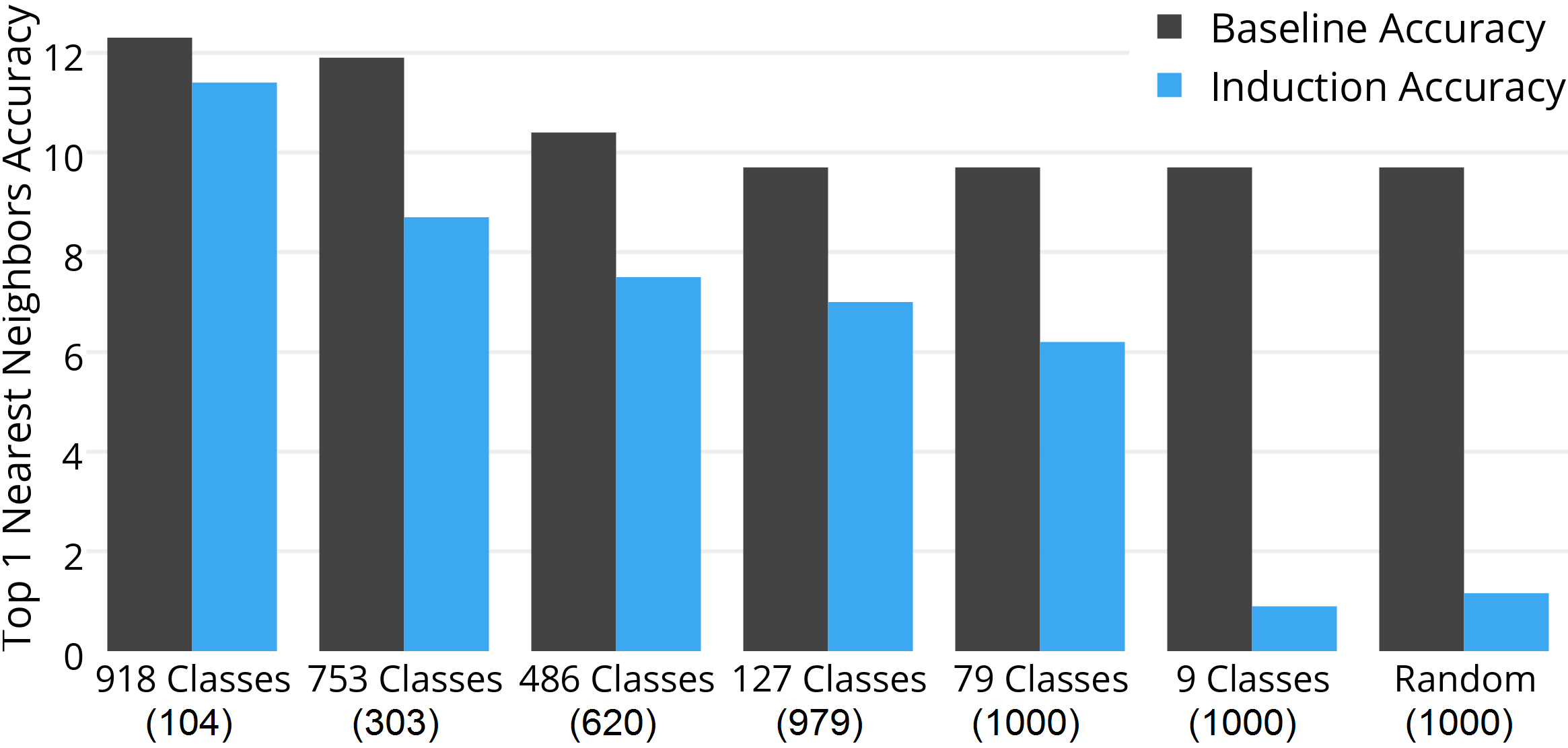}
\includegraphics[width=0.495\linewidth]{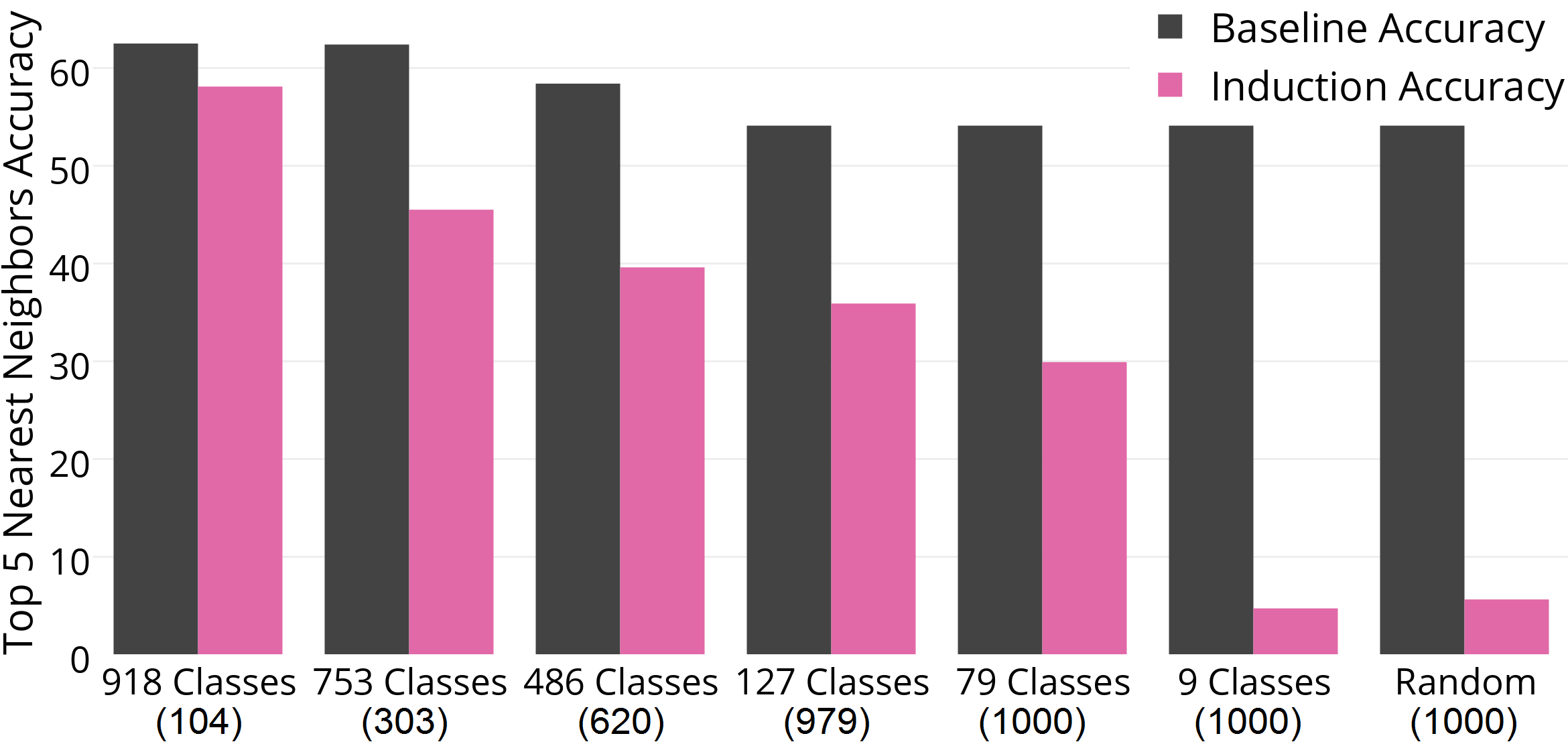}
\end{center}
\vspace{-.1in}
\caption{Does a CNN trained for discriminating between coarse classes learns a feature embedding capable of distinguishing between fine classes? We quantified this by measuring the induction accuracy defined as following: after training a feature embedding for a particular set of classes (set A), the induction accuracy is the nearest neighbor (top-1 and top-5) classification accuracy measured in the FC8 feature space of the subset of 1000 ImageNet classes not present in set A. The syntax on the x-axis\;{\tt A Classes(B)} indicates that the network was trained with A classes and the induction accuracy was measured on B classes. The baseline accuracy is the accuracy on B classes when the CNN was trained for all 1000 classes. The margin between the baseline and the induction accuracy indicates a drop in the network's ability to distinguish fine classes when being trained on coarse classes. The results show that features learnt by pre-training on just 127 classes still lead to fairly good induction.}
\vspace{-.2in}
\label{fig:induction}
\end{figure*}

\section{How does the amount of pre-training data affect transfer performance?}
\label{sec:examples_per_class}

For answering this question, we trained 5 different AlexNet models from scratch using 50, 125, 250, 500 and 1000 images per each of the 1000 ImageNet classes using the procedure described in Section~\ref{sec:setup}. The variation in performance with amount of pre-training data when these models are finetuned for PASCAL-DET, PASCAL-ACT-CLS and SUN-CLS is shown in Figure~\ref{fig:number-of-class}. For PASCAL-DET, the mean average precision (mAP) for CNNs with 1000, 500 and 250 images/class is found to be 58.3, 57.0 and 54.6. A similar trend is observed for PASCAL-ACT-CLS and SUN-CLS. 
%it is found to be 59.8, 56.5 and 51.4, and lastly the class accuracy for SUN is found to be 52.2, 49.7, and 46.7 respectively. 
These results indicate that using half the amount of pre-training data leads to only a \textit{marginal} reduction in performance on transfer tasks. It is important to note that the performance on the ImageNet classification task (the pre-training task) steadily increases with the amount of training data, whereas on transfer tasks, the performance increase with respect to additional pre-training data is significantly slower. This suggests that while adding additional examples to ImageNet classes will improve the ImageNet performance, it has diminishing return for transfer task performance.

\section{How does the taxonomy of the pre-training task affect transfer performance?}
In the previous section we investigated how varying number of pre-training images per class effects the performance in transfer tasks. Here we investigate the flip side: keeping the amount of data constant while changing the nomenclature of training labels. 
%Through manipulating different aspects of the ImageNet dataset, we hope to narrow in on what makes ImageNet good for transfer tasks.

\begin{figure*}[t!]
\begin{center}
\includegraphics[width=1.0\linewidth]{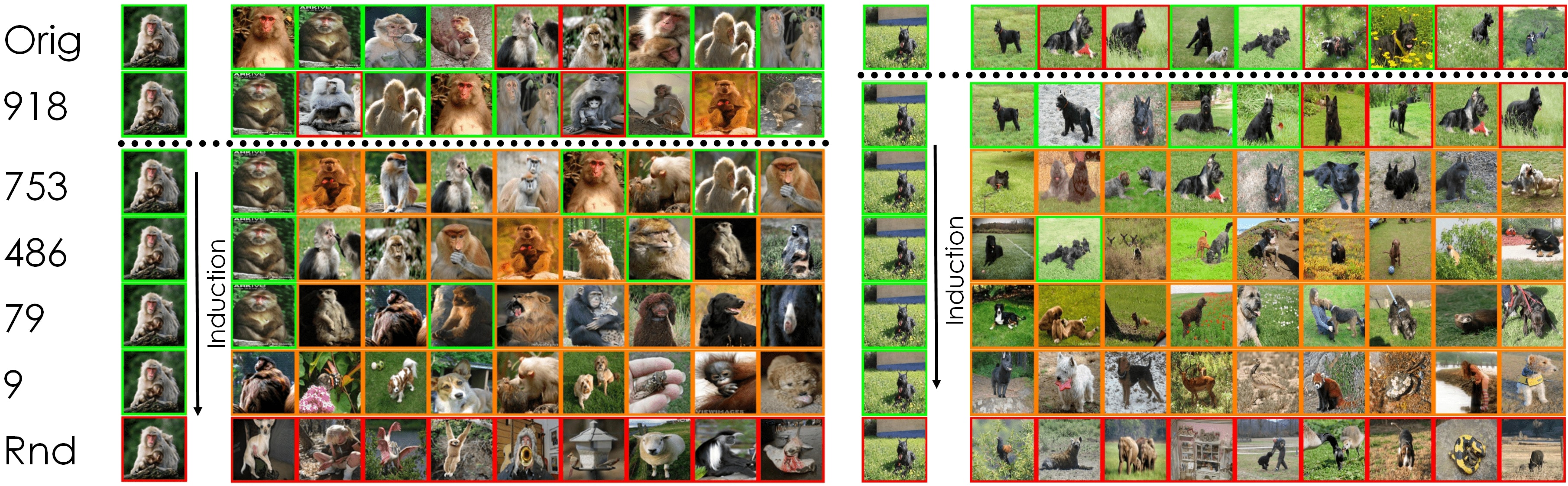}
\end{center}
\vspace{-.2in}
\caption{Can feature embeddings obtained by training on coarse classes be able to distinguish fine classes they were never trained on? E.g. by training on monkeys, can the network pick out macaques? Here we look at the FC7 nearest neighbors (NN) of two randomly sampled images: a macaque (left column) and a giant schnauzer (right column), with each row showing feature embeddings trained with different number of classes (from fine to coarse). The row(s) above the dotted line indicate that the image class (i.e. macaque/giant schnauzer) was one of the training classes, whereas in rows below the image class was not present in the training set. Images in green indicate that the NN image belongs to the correct fine class (i.e. either macaque or giant schnauzer); orange indicates the correct coarse class (based on the WordNet hierarchy) but incorrect fine class; red indicated incorrect coarse class.  All green images below the dotted line indicate instances of correct fine-grain nearest neighbor retrieval for features that were never trained on that class.}
\vspace{-.1in}
\label{fig:nn}
\end{figure*}

\subsection{The effect of number of pre-training classes on transfer performance}
\label{sec:finegrain}
The 1000 classes of the ImageNet challenge \cite{ILSVRC15} are derived from leaves of the WordNet tree \cite{WordNet}. Using this tree, it is possible to generate different class taxonomies while keeping the total number of images constant. One can generate taxonomies in two ways: (1) bottom up clustering, wherein the leaf nodes belonging to a common parent are iteratively clustered together (see Figure~\ref{fig:imgnttree}), or (2) by fixing the distance of the nodes from the root node (i.e. top down clustering). Using bottom up clustering, 18 possible taxonomies can be generated. Among these, we chose 5 sets of labels constituting 918, 753, 486, 79 and 9 classes respectively. Using top-down clustering only 3 label sets of 127, 10 and 2 can be generated, and we used the one with 127 classes. For studying the effect of number of pre-training classes on transfer performance, we trained separate AlexNet CNNs from scratch using these label sets. 

Figure~\ref{fig:hierarchy} shows the effect of number of pre-training classes obtained using bottom up clustering of WordNet tree on transfer performance. We also include the performance of these different networks on the Imagenet classification task itself after finetuning only the last layer to distinguish between all the 1000 classes. The results show that increase in performance on transfer tasks is significantly slower with increase in number of classes as compared to performance on Imagenet itself. Using only 486 classes results in a performance drop of 1.7 mAP for PASCAL-DET, 0.8\% accuracy for SUN-CLS and a boost of 0.6 mAP for PASCAL-ACT-CLS. Table \ref{table:coarse} shows the transfer performance after pre-training with 127 classes obtained from top down clustering. The results from this table and the figure indicate that only diminishing returns in transfer performance are observed when more than 127 classes are used. Our results also indicate that making the ImageNet classes finer will not help improve transfer performance.

It can be argued that the PASCAL task requires discrimination between only 20 classes and therefore pre-training with only 127 classes should not lead to substantial reduction in performance. However, the trend also holds true for SUN-CLS that requires discrimination between 397 classes. These two results taken together suggest that although training with a large number of classes is beneficial, diminishing returns are observed beyond using 127 distinct classes for pre-training.

Furthermore, for PASCAL-ACT-CLS and SUN-CLS, finetuning on CNNs pre-trained with class set sizes of 918, and 753 actually results in better performance than using all 1000 classes. This may indicate that having too many classes for pre-training works against learning good generalizable features. Hence, when generating a dataset, one should be attentive of the nomenclature of the classes.

\subsection{Is fine-grain recognition necessary for learning transferable features?}
\label{sec:fine-necessary}
ImageNet challenge requires a classifier to distinguish between 1000 classes, some of which are very fine-grained, such as different breeds of dogs and cats.  Indeed, most humans do not perform well on ImageNet unless specifically trained~\cite{ILSVRC15}, and yet are easily able to perform most everyday visual tasks.  This raises the question: is fine-grained recognition necessary for CNN models to learn good feature representations, or is coarse-grained object recognition (e.g. just distinguishing cats from dogs) is sufficient?

Note that the label set of 127 classes from the previous experiment contains 65 classes that are present in the original set of 1000 classes and the remainder are inner nodes of the WordNet tree. However, all these 127 classes (see supplementary materials) represent coarse semantic concepts.  As discussed earlier, pre-training with these classes results in only a small drop in transfer performance (see Table \ref{table:coarse}). This suggests that performing fine-grained recognition is only marginally helpful and does not appear to be critical for learning good transferable features.

\subsection{Does training with coarse classes induce features relevant for fine-grained recognition?}
\label{sec:relevant}
Earlier, we have shown that the features learned on the 127 coarse classes perform almost as well on our transfer tasks as the full set of 1000 ImageNet classes.  Here we will probe this further by asking a different question: is the feature embedding induced by the coarse class classification task capable of separating the fine labels of ImageNet (which it never saw at training)?  

To investigate this, we used top-1 and top-5 nearest neighbors in the FC7 feature space to measure the accuracy of identifying fine-grained ImageNet classes after training only on a set of coarse classes. We call this measure, ``induction accuracy''. As a qualitative example, Figure~\ref{fig:nn} shows nearest neighbors for a macaque (left) and a schnauzer (right) for feature embeddings trained on ImageNet but with different number of classes. All green-border images below the dotted line indicate instances of correct fine-grain nearest neighbor retrieval for features that were never trained on that class.  

Quantitative results are shown in Figure~\ref{fig:induction}. The results show that when 127 classes are used, fine-grained recognition k-NN performance is only about 15\% lower compared to training directly for these fine-grained classes (i.e. baseline accuracy). This is rather surprising and suggests that CNNs implicitly discover features capable of distinguishing between finer classes while attempting to distinguish between relatively coarse classes. 

\begin{figure}[t]
\begin{center}
\includegraphics[width=1.0\linewidth]{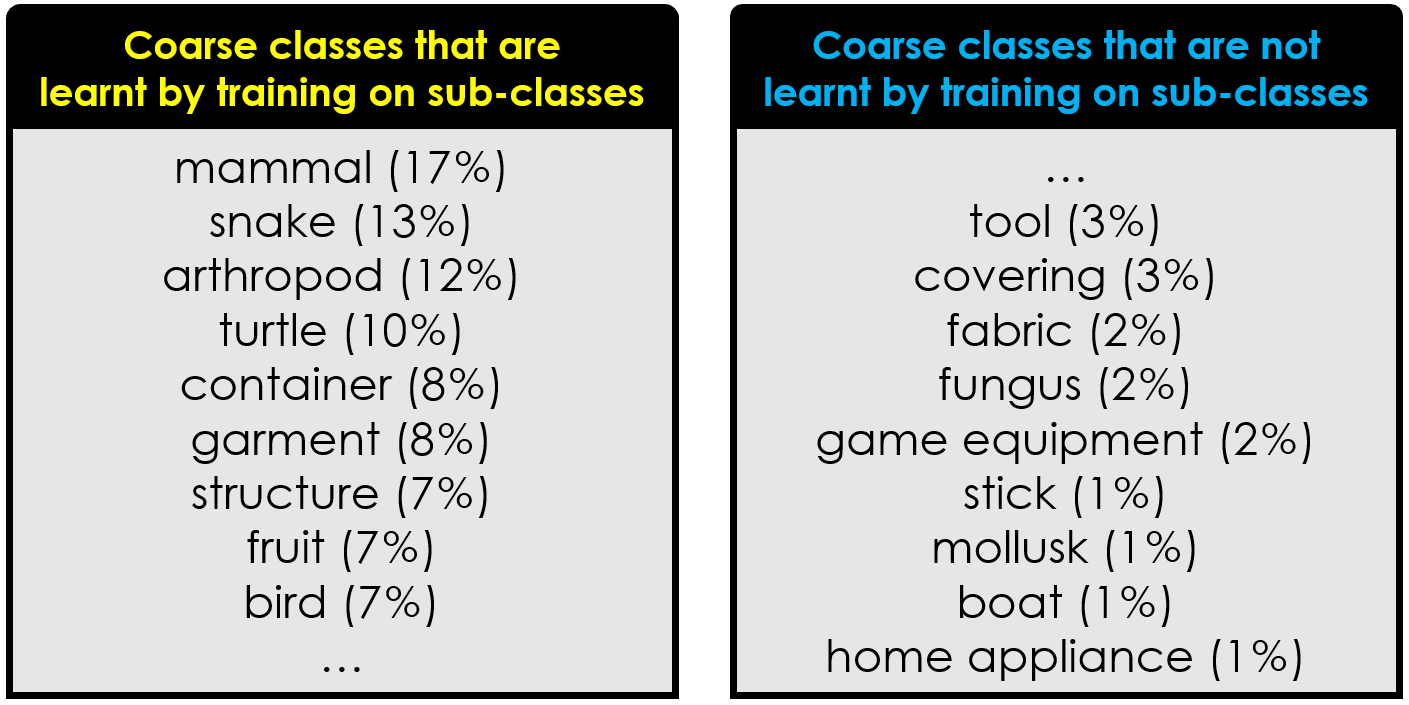}
\end{center}
\vspace{-.2in}
\caption{Does the network learn to discriminate coarse semantic concepts by training only on finer sub-classes? The degree to which the concept of coarse class is learnt was quantified by measuring the difference (in percentage points) between the accuracy of classifying the coarse class and the average accuracy of individually classifying all the sub-classes of this coarse class. Here, the top and bottom classes sorted by this metric are shown using the label set of size 127 with classes with at least 5 subclasses. We observe that classes whose subclasses are visually consistent (e.g. mammal) are better represented than these that are visually dissimilar (e.g. home appliance). 
}
\label{fig:f_2_c}
\end{figure}

\subsection{Does training with fine-grained classes induce features relevant for coarse recognition?}
\label{sec:relevant2}
Investigating whether the network learns features relevant for fine-grained recognition by training on coarse classes raises the reverse question: does training with fine-grained classes induce features relevant for coarse recognition? If this is indeed the case, then we would expect that when a CNN makes an error, it is more likely to confuse a sub-class (i.e. error in fine-grained recognition) with other sub-classes of the same coarse class. This effect can be measured by computing the difference between the accuracy of classifying the coarse class and the average accuracy of individually classifying all the sub-classes of this coarse class (please see supplementary materials for details). 

Figure~\ref{fig:f_2_c} shows the results. We find that coarse semantic classes such as mammal, fruit, bird, etc. that contain visually similar sub-classes show the hypothesized effect, whereas classes such as tool and home appliance that contain visually dissimilar subclasses do not exhibit this effect. These results indicate that subclasses that share a common visual structure allow the CNN to learn features that are more generalizable.  This might suggest a way to improve feature generalization by making class labels respect visual commonality rather than simply WordNet semantics. 

\subsection{More Classes or More Examples Per Class?}
\label{sec:mc_or_me}

Results in previous sections show that it is possible to achieve good performance on transfer tasks using significantly less pre-training data and fewer pre-training classes. However it is unclear what is more important -- the number of classes or the number or examples per class. One extreme is to only have 1 class and all 1.2M images from this class and the other extreme is to have 1.2M classes and 1 image per class. It is clear that both ways of splitting the data will result in poor generalization, so the answer must lie somewhere in-between.
%and there exists an optimal way of splitting the data into classes.

To investigate this, we split the same amount of pre-training data in two ways: (1) more classes with fewer images per class, and (2) fewer classes with more images per class. We use datasets of size 500K, 250K and 125K images for this experiment. For 500K images, we considered two ways of constructing the training set -- (1) 1000 classes with 500 images/class, and (2) 500 classes with 1000 images/class. Similar splits were made for data budgets of 250K and 125K images. The 500, 250 and 125 classes for these experiments were drawn from a uniform distribution among the 1000 ImageNet classes. Similarly, the image subsets containing 500, 250 and 125 images were drawn from a uniform distribution among the images that belong to the class.

The results presented in Table~\ref{table:class-vs-example} show that having more images per class with fewer number of classes results in features that perform very slightly better on PASCAL-DET, whereas for SUN-CLS, the performance is comparable across the two settings.

\setlength{\tabcolsep}{4pt}
\begin{table}[t]
\begin{center}
\scalebox{0.805}{
    \begin{tabular}{l|ccc|ccc}
    \hline
    Dataset & & PASCAL & &  & SUN &\\
    \hline
    Data size & 500K & 250K & 125K & 500K & 250K & 125K \\
    \hline
    More examples/class  & 57.1 & 54.8 & 50.6  & 50.6 & 45.7 & 42.2 \\
    More classes & 57.0 & 52.5 & 49.8  & 49.7 & 46.7 & 42.3 \\
    \hline
    \end{tabular}
}
\vspace{-.1in}
\caption{For a fixed budget of pre-training data, is it better to have more examples per class and fewer classes or vice-versa? The row `more examples/class` was pretrained with subsets of ImageNet containing 500, 250 and 125 classes with 1000 examples each. The row `more classes` was pretrained with 1000 classes, but 500, 250 and 125 examples each.  Interestingly, the transfer performance on both PASCAL and SUN appears to be broadly similar under both scenarios. 
}
\vspace{-.1in}
\label{table:class-vs-example}
\end{center}
\end{table}
\setlength{\tabcolsep}{1.4pt}
\normalsize

\setlength{\tabcolsep}{4pt}
\begin{table}[t]
\begin{center}
\begin{tabular}{l|ccc}
\hline
\; Pre-trained Dataset \; & \qquad PASCAL \qquad \qquad&\\
\hline
\; ImageNet \; & \qquad 58.3  $\pm$ \; 0.3 \qquad \qquad&\\
\; Pascal removed ImageNet \;& \qquad 57.8 $\pm$ \; 0.1 \qquad \qquad&\\
\; Places \;& \qquad 53.8 $\pm$ \; 0.1 \qquad \qquad&\\
\hline
\end{tabular}
\vspace{-.1in}
\caption{PASCAL-DET results after pre-training on entire ImageNet, PASCAL-removed-ImageNet and Places data sets. Removing PASCAL classes from ImageNet leads to an insignificant reduction in performance.}
\vspace{-.1in}
\label{table:pascal-removed}
\end{center}
\end{table}
\setlength{\tabcolsep}{1.4pt}

\subsection{How important is to pre-train on classes that are also present in a target task?}
It is natural to expect that higher correlation between pre-training and transfer tasks leads to better performance on a transfer task. This indeed has been shown to be true in \cite{yosinski2014transferable}. One possible source of correlation between pre-training and transfer tasks are classes common to both tasks. In order to investigate how strong is the influence of these common classes, we ran an experiment where we removed all the classes from ImageNet that are contained in the PASCAL challenge. PASCAL has 20 classes, some of which map to more than one ImageNet class and thus, after applying this exclusion criterion we are only left with 771 ImageNet classes. 

\begin{figure}[t]
    \begin{center}
    \includegraphics[width=1.0\linewidth]{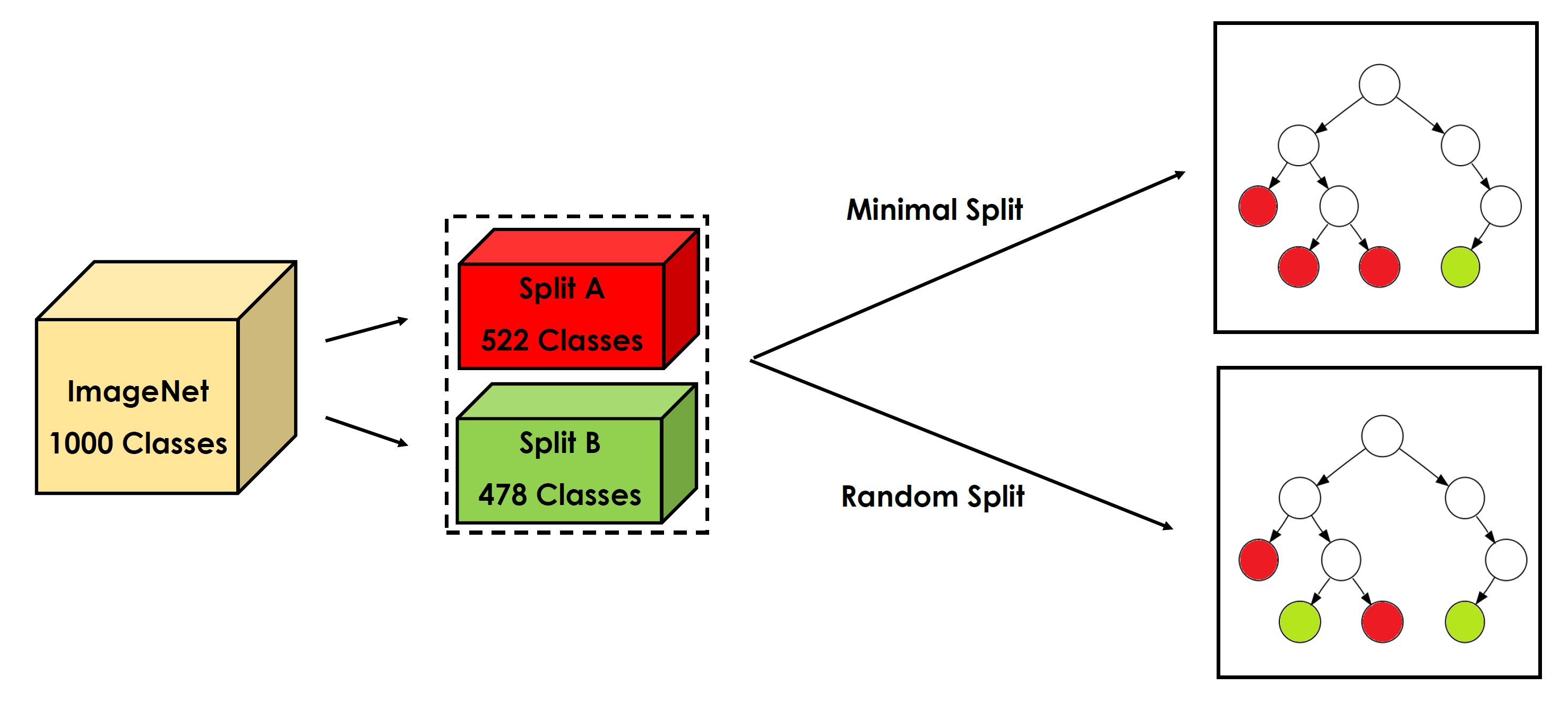}\\
    \end{center}
    \vspace{-.2in}
    \caption{An illustration of the procedure used to split the ImageNet dataset. Splits were constructed in 2 different ways. The random split selects classes at random from the 1000 ImageNet classes. The minimal split is made in a manner that ensures no two classes in the same split have a common ancestor up to depth four of WordNet tree. Collage in Figure~\ref{fig:split-col} visualizes the random and minimal splits.}
%    \vspace{-.1in}
    \label{fig:split-vis}
\end{figure}

Table \ref{table:pascal-removed} compares the results on PASCAL-DET when the PASCAL-removed-ImageNet is used for pre-training against the original ImageNet and a baseline of pre-training on the Places~\cite{placesdata} dataset. The PASCAL-removed-ImageNet achieves mAP of 57.8 (compared to 58.3 with the full ImageNet) indicating that training on ImageNet classes that are not present in PASCAL is sufficient to learn features that are also good for PASCAL classes. 

\section{Does data augmentation from non-target classes always improve performance?}
\label{sec:more_class}

The analysis using PASCAL-removed ImageNet indicates that pre-training on non-PASCAL classes aids performance on PASCAL. This raises the question: is it always better to add pre-training data from additional classes that are not part of the target task? To investigate and test this hypothesis, we chose two different methods of splitting the ImageNet classes. The first is random split, in which the 1000 ImageNet classes are split randomly; the second is a minimal split, in which the classes are deliberately split to ensure that similar classes are not in the same split, (Figure~\ref{fig:split-vis}). In order to determine if additional data helps performance for classes in split A, we pre-trained two CNNs -- one for classifying all classes in split A and the other for classifying all classes in both split A and B (i.e. full dataset). We then finetuned the last layer of the network trained on the full dataset on split A only. If it is the case that additional data from split B helps performance on split A, then the CNN pre-trained with the full dataset should perform better than CNN pre-trained only on split A. 

\begin{figure}[t!]
    \begin{center}
    \includegraphics[width=1.0\linewidth]{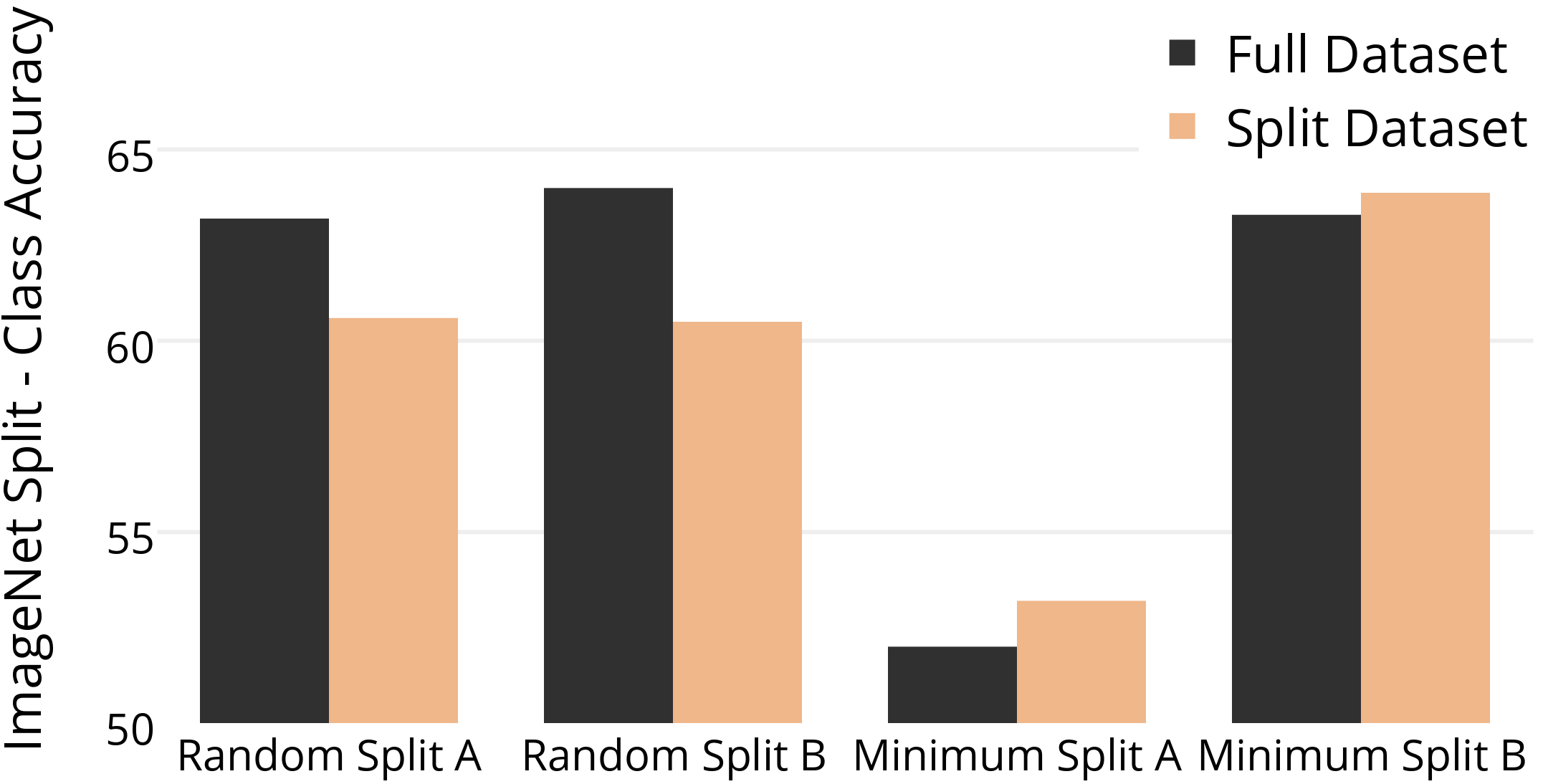}
    \end{center}
    \vspace{-.2in}
    \caption{Does adding arbitrary classes to pre-training data always improve transfer performance? This question was tested by training two CNNs, one for classifying classes in split A and other for classifying classes in split A and B both. We then finetuned the CNN trained on both the splits on split A. If it is the case that adding more pre-training data helps, then performance of the CNN pre-trained on both the splits (black) should be higher than a CNN pre-trained on a single split (orange). For random splits, this indeed is the case, whereas for minimal splits adding more pre-training data hurts performance. This suggests, that additional pre-training data is useful only if it is correlated to the target task.}
    \label{figure:split}
\end{figure}

Using the random split, Figure~\ref{figure:split} shows that the results of this experiment confirms the intuition that additional data is indeed useful for both splits. However, under a random class split within ImageNet, we are almost certain to have extremely similar classes (e.g. two different breeds of dogs) ending up on the different sides of the split. So, what we have shown so far is that we can improve performance on, say, husky classification by also training on poodles. Hence, the motivation for the minimal split: does adding arbitrary, unrelated classes, such as fire trucks, help dog classification?

\begin{figure}[t]
\begin{center}
\includegraphics[width=1.0\linewidth]{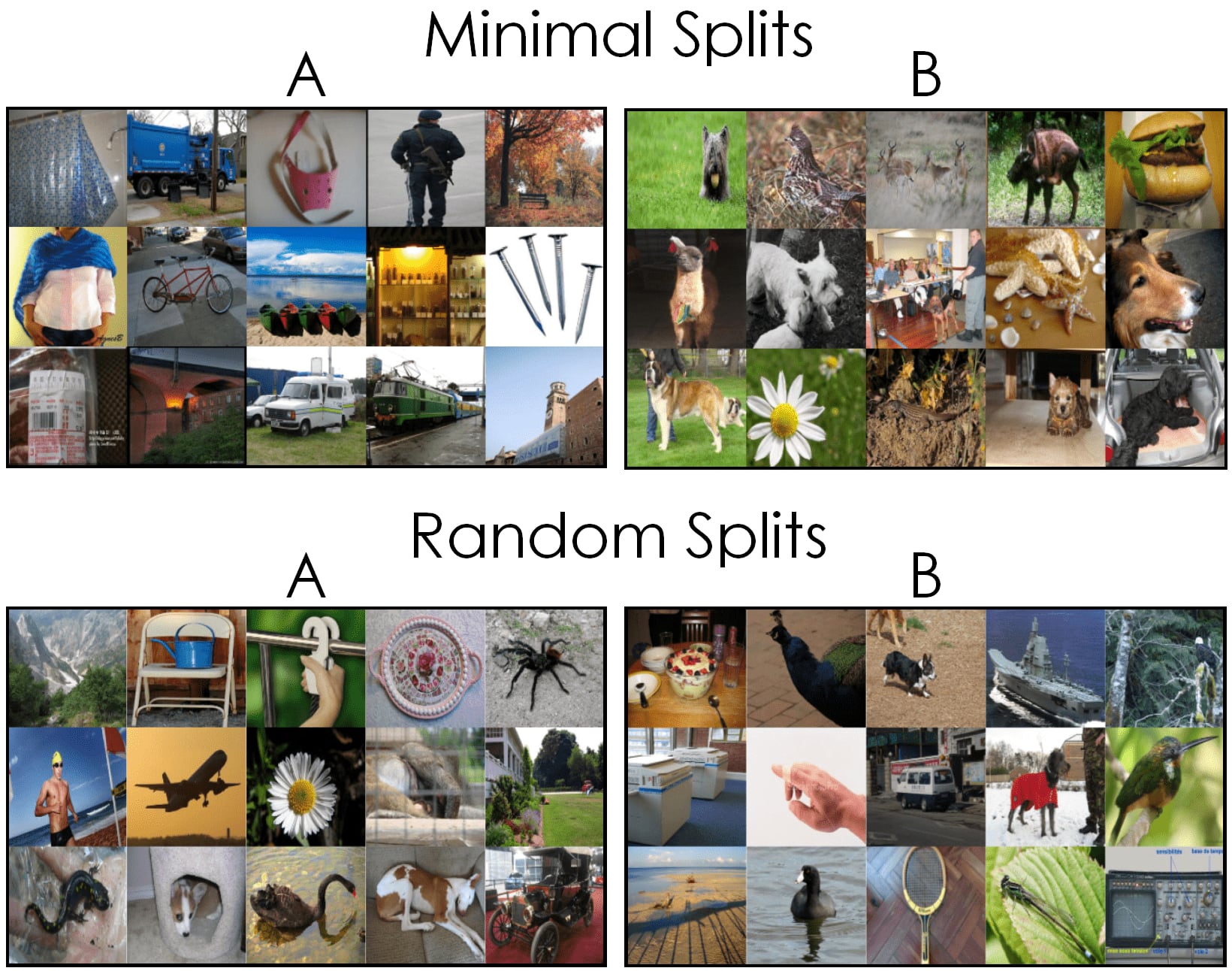}
\end{center}
\vspace{-.2in}
\caption{Visualization of the random and minimal splits used for testing - is adding more pre-training data always useful? The two minimal sets contain disparate sets of objects. The minimal split A and B consists mostly of inanimate objects and living things respectively. On the other hand, random splits contain semantically similar objects.}
\label{fig:split-col}
\end{figure}

The classes in minimal split A do not share any common ancestor with minimal split B up until the nodes at depth 4 of the WordNet hierarchy (Figure~\ref{fig:split-vis}). This ensures that any class in split A is sufficiently disjoint from split B. Split A has 522 classes and split B has 478 classes (N.B.: for consistency, random splits A and B also had the same number of classes). In order to intuitively understand the difference between min splits A and B, we have visualized a random sample of images in these splits in Figure~\ref{fig:split-col}. Min split A consists of mostly static images and min split B consists of living objects.

Contrary to the earlier observation, Figure~\ref{figure:split} shows that both min split A and B performs better than the full dataset when we finetune only the last layer. This result is quite surprising because it shows that finetuning the last layer from a network pre-trained on the full dataset, it is not possible to match the performance of a network trained on just one split. We have observed that when training all the layers for an extensive amount of time (420K iterations), the accuracy of min split A does benefit from pre-training on split B but does \textit{not} for min split B. One explanation could be that images in split B (e.g. person) is contained in images in split A, (e.g. buildings, clothing) but \textit{not} vice versa.

While it might be possible to recover performance with very clever adjustments of learning rates, current results suggest that training with data from unrelated classes may push the network into a local minimum from which it might be hard to find a better optima that can be obtained by training the network from scratch. 

\section{Discussion}

In this work we analyzed factors that affect the quality of ImageNet pre-trained features for transfer learning.
Our goal was not to consider alternative neural network architectures, but rather to establish facts about which aspects of the {\em training data} are important for feature learning.  

The current consensus in the field is that the key to learning highly generalizable deep features is the large amounts of training data and the large number of classes.

To quote the influential R-CNN paper: \textit{``..success resulted from training a large CNN on 1.2 million labeled images...''}~\cite{girshick2014rich}. After the publication of R-CNN, most researchers assumed that the full ImageNet is necessary to pre-train good general-purpose features.  Our work quantitatively questions this assumption, and yields some quite surprising results. For example, we have found that a significant reduction in the number of classes or the number of images used in pre-training has only a modest effect on transfer task performance.  

While we do not have an explanation as to the cause of this resilience, we list some speculative possibilities that should inform further study of this topic:

\begin{itemize}[leftmargin=*]
\item In our experiments, we investigated only one CNN architecture -- AlexNet. While ImageNet-trained AlexNet features are currently the most popular starting point for fine-tuning on transfer tasks, there exist deeper architectures such as VGG~\cite{simonyan2014vgg},  ResNet~\cite{he2015resnet}, and GoogLeNet~\cite{Szegedy2015}. It would be interesting to see if our findings hold up on deeper networks. If not, it might suggest that AlexNet capacity is less than previously thought.
\item Our results might indicate that researchers have been overestimating the amount of data required for learning good general CNN features. If that is the case, it might suggest that CNN training is not as data-hungry as previously thought.  It would also suggest that beating ImageNet-trained features with models trained on a much bigger data corpus will be much harder than once thought.
\item Finally, it might be that the currently popular target tasks, such as PASCAL and SUN, are too similar to the original ImageNet task to really test the generalization of the learned features. Alternatively, perhaps a more appropriate approach to test the generalization is with much less fine-tuning (e.g. one-shot-learning) or no fine-tuning at all (e.g. nearest neighbour in the learned feature space).  
\end{itemize}

In conclusion, while the answer to the titular question ``What makes ImageNet good for transfer learning?'' still lacks a definitive answer, our results have shown that a lot of ``folk wisdom'' on why ImageNet works well is not accurate.  We hope that this paper will pique our colleagues' curiosity and facilitate further research on this fascinating topic. 

\section{Acknowledgements}
This work was supported in part by ONR MURI N00014-14-1-0671. We gratefully acknowledge NVIDIA corporation for the donation of K40 GPUs and access to the NVIDIA PSG cluster for this research. We would like to acknowledge the support from the Berkeley Vision and Learning Center (BVLC) and Berkeley DeepDrive (BDD). Minyoung Huh was partially supported by the Rose Hill Foundation. 

\small
\bibliographystyle{ieee}
\bibliography{all-refs}

\begin{thebibliography}{10}

\bibitem{agrawal2015learning}
P.~Agrawal, J.~Carreira, and J.~Malik.
\newblock Learning to see by moving.
\newblock In {\em Proceedings of the IEEE International Conference on Computer
  Vision}, pages 37--45, 2015.

\bibitem{agrawal2014analyzing}
P.~Agrawal, R.~Girshick, and J.~Malik.
\newblock Analyzing the performance of multilayer neural networks for object
  recognition.
\newblock In {\em Computer Vision--ECCV 2014}, pages 329--344. Springer, 2014.

\bibitem{azizpour2015generic}
H.~Azizpour, A.~Razavian, J.~Sullivan, A.~Maki, and S.~Carlsson.
\newblock From generic to specific deep representations for visual recognition.
\newblock In {\em Proceedings of the IEEE Conference on Computer Vision and
  Pattern Recognition Workshops}, pages 36--45, 2015.

\bibitem{bengio2012unsupervised}
Y.~Bengio, A.~C. Courville, and P.~Vincent.
\newblock Unsupervised feature learning and deep learning: A review and new
  perspectives.
\newblock {\em CoRR, abs/1206.5538}, 1, 2012.

\bibitem{bourlard1988auto}
H.~Bourlard and Y.~Kamp.
\newblock Auto-association by multilayer perceptrons and singular value
  decomposition.
\newblock {\em Biological cybernetics}, 59(4-5):291--294, 1988.

\bibitem{carreira2015human}
J.~Carreira, P.~Agrawal, K.~Fragkiadaki, and J.~Malik.
\newblock Human pose estimation with iterative error feedback.
\newblock {\em arXiv preprint arXiv:1507.06550}, 2015.

\bibitem{dai2015instance}
J.~Dai, K.~He, and J.~Sun.
\newblock Instance-aware semantic segmentation via multi-task network cascades.
\newblock {\em arXiv preprint arXiv:1512.04412}, 2015.

\bibitem{doersch2015unsupervised}
C.~Doersch, A.~Gupta, and A.~A. Efros.
\newblock Unsupervised visual representation learning by context prediction.
\newblock In {\em Proceedings of the IEEE International Conference on Computer
  Vision}, pages 1422--1430, 2015.

\bibitem{donahue2015long}
J.~Donahue, L.~Anne~Hendricks, S.~Guadarrama, M.~Rohrbach, S.~Venugopalan,
  K.~Saenko, and T.~Darrell.
\newblock Long-term recurrent convolutional networks for visual recognition and
  description.
\newblock In {\em Proceedings of the IEEE Conference on Computer Vision and
  Pattern Recognition}, pages 2625--2634, 2015.

\bibitem{donahue2013decaf}
J.~Donahue, Y.~Jia, O.~Vinyals, J.~Hoffman, N.~Zhang, E.~Tzeng, and T.~Darrell.
\newblock Decaf: A deep convolutional activation feature for generic visual
  recognition.
\newblock {\em arXiv preprint arXiv:1310.1531}, 2013.

\bibitem{WordNet}
C.~Fellbaum.
\newblock {\em WordNet: An Electronic Lexical Database}.
\newblock Bradford Books, 1998.

\bibitem{girshick2014rich}
R.~Girshick, J.~Donahue, T.~Darrell, and J.~Malik.
\newblock Rich feature hierarchies for accurate object detection and semantic
  segmentation.
\newblock In {\em Computer Vision and Pattern Recognition (CVPR), 2014 IEEE
  Conference on}, pages 580--587. IEEE, 2014.

\bibitem{rstarcnn2015}
G.~Gkioxari, R.~Girshick, and J.~Malik.
\newblock Contextual action recognition with r\*cnn.
\newblock In {\em ICCV}, 2015.

\bibitem{goroshin2015unsupervised}
R.~Goroshin, J.~Bruna, J.~Tompson, D.~Eigen, and Y.~LeCun.
\newblock Unsupervised feature learning from temporal data.
\newblock {\em arXiv preprint arXiv:1504.02518}, 2015.

\bibitem{he2015resnet}
K.~He, X.~Zhang, S.~Ren, and J.~Sun.
\newblock Deep residual learning for image recognition.
\newblock {\em CoRR}, abs/1512.03385, 2015.

\bibitem{jayaraman2015learning}
D.~Jayaraman and K.~Grauman.
\newblock Learning image representations tied to ego-motion.
\newblock In {\em Proceedings of the IEEE International Conference on Computer
  Vision}, pages 1413--1421, 2015.

\bibitem{caffe}
Y.~Jia.
\newblock {Caffe}: An open source convolutional architecture for fast feature
  embedding.
\newblock \url{http://caffe.berkeleyvision.org/}, 2013.

\bibitem{armand2015weak}
A.~Joulin, L.~van~der Maaten, A.~Jabri, and N.~Vasilache.
\newblock Learning visual features from large weakly supervised data.
\newblock In {\em ECCV}, 2016.

\bibitem{karpathy2015deep}
A.~Karpathy and L.~Fei-Fei.
\newblock Deep visual-semantic alignments for generating image descriptions.
\newblock In {\em Proceedings of the IEEE Conference on Computer Vision and
  Pattern Recognition}, pages 3128--3137, 2015.

\bibitem{kingma2013auto}
D.~P. Kingma and M.~Welling.
\newblock Auto-encoding variational bayes.
\newblock {\em arXiv preprint arXiv:1312.6114}, 2013.

\bibitem{philipp2016init}
P.~Kr\"ahenb\"uhl, C.~Doersch, J.~Donahue, and T.~Darrell.
\newblock Data-dependent initializations of convolutional neural networks.
\newblock In {\em ICLR}, 2016.

\bibitem{krizhevsky2012ImageNet}
A.~Krizhevsky, I.~Sutskever, and G.~E. Hinton.
\newblock Imagenet classification with deep convolutional neural networks.
\newblock In {\em Advances in neural information processing systems}, pages
  1097--1105, 2012.

\bibitem{larsson2016learning}
G.~Larsson, M.~Maire, and G.~Shakhnarovich.
\newblock Learning representations for automatic colorization.
\newblock In {\em ECCV}, 2016.

\bibitem{lecun2015deep}
Y.~LeCun, Y.~Bengio, and G.~Hinton.
\newblock Deep learning.
\newblock {\em Nature}, 521(7553):436--444, 2015.

\bibitem{lecun1989backpropagation}
Y.~LeCun, B.~Boser, J.~S. Denker, D.~Henderson, R.~E. Howard, W.~Hubbard, and
  L.~D. Jackel.
\newblock Backpropagation applied to handwritten zip code recognition.
\newblock {\em Neural computation}, 1(4):541--551, 1989.

\bibitem{li2016withoutforgetting}
Z.~Li and D.~Hoiem.
\newblock Learning without forgetting.
\newblock In {\em ECCV}, 2016.

\bibitem{mobahi2009deep}
H.~Mobahi, R.~Collobert, and J.~Weston.
\newblock Deep learning from temporal coherence in video.
\newblock In {\em Proceedings of the 26th Annual International Conference on
  Machine Learning}, pages 737--744. ACM, 2009.

\bibitem{noroozi2016puzzles}
M.~Noroozi and F.~Paolo.
\newblock Unsupervised learning of visual representations by solving jigsaw
  puzzles.
\newblock In {\em ECCV}, 2016.

\bibitem{olshausen1996emergence}
B.~A. Olshausen et~al.
\newblock Emergence of simple-cell receptive field properties by learning a
  sparse code for natural images.
\newblock {\em Nature}, 381(6583):607--609, 1996.

\bibitem{owens2016sound}
A.~Owens, P.~Isola, J.~McDermott, A.~Torralba, E.~Adelson, and F.~William.
\newblock Visually indicated sounds.
\newblock In {\em CVPR}, 2016.

\bibitem{pathak2016context}
D.~Pathak, P.~Kr\"ahenb\"uhl, J.~Donahue, T.~Darrell, and A.~Efros.
\newblock Context encoders: Feature learning by inpainting.
\newblock In {\em CVPR}, 2016.

\bibitem{ranzato2007unsupervised}
M.~Ranzato, F.~J. Huang, Y.-L. Boureau, and Y.~LeCun.
\newblock Unsupervised learning of invariant feature hierarchies with
  applications to object recognition.
\newblock In {\em Computer Vision and Pattern Recognition, 2007. CVPR'07. IEEE
  Conference on}, pages 1--8. IEEE, 2007.

\bibitem{razavian2014cnn}
A.~Razavian, H.~Azizpour, J.~Sullivan, and S.~Carlsson.
\newblock Cnn features off-the-shelf: an astounding baseline for recognition.
\newblock In {\em Proceedings of the IEEE Conference on Computer Vision and
  Pattern Recognition Workshops}, pages 806--813, 2014.

\bibitem{ren2015faster}
S.~Ren, K.~He, R.~Girshick, and J.~Sun.
\newblock Faster r-cnn: Towards real-time object detection with region proposal
  networks.
\newblock In {\em Advances in Neural Information Processing Systems}, pages
  91--99, 2015.

\bibitem{ILSVRC15}
O.~Russakovsky, J.~Deng, H.~Su, J.~Krause, S.~Satheesh, S.~Ma, Z.~Huang,
  A.~Karpathy, A.~Khosla, M.~Bernstein, A.~C. Berg, and L.~Fei-Fei.
\newblock {ImageNet Large Scale Visual Recognition Challenge}.
\newblock {\em International Journal of Computer Vision (IJCV)}, 2015.

\bibitem{salakhutdinov2009deep}
R.~Salakhutdinov and G.~E. Hinton.
\newblock Deep boltzmann machines.
\newblock In {\em International Conference on Artificial Intelligence and
  Statistics}, pages 448--455, 2009.

\bibitem{sermanet2013overfeat}
P.~Sermanet, D.~Eigen, X.~Zhang, M.~Mathieu, R.~Fergus, and Y.~LeCun.
\newblock Overfeat: Integrated recognition, localization and detection using
  convolutional networks.
\newblock {\em arXiv preprint arXiv:1312.6229}, 2013.

\bibitem{simonyan2014two}
K.~Simonyan and A.~Zisserman.
\newblock Two-stream convolutional networks for action recognition in videos.
\newblock In {\em Advances in Neural Information Processing Systems}, pages
  568--576, 2014.

\bibitem{simonyan2014vgg}
K.~Simonyan and A.~Zisserman.
\newblock Very deep convolutional networks for large-scale image recognition.
\newblock {\em CoRR}, abs/1409.1556, 2014.

\bibitem{Szegedy2015}
C.~Szegedy, W.~Liu, Y.~Jia, P.~Sermanet, S.~Reed, D.~Anguelov, D.~Erhan,
  V.~Vanhoucke, and A.~Rabinovich.
\newblock Going deeper with convolutions.
\newblock In {\em CVPR}, 2015.

\bibitem{wang2015unsupervised}
X.~Wang and A.~Gupta.
\newblock Unsupervised learning of visual representations using videos.
\newblock In {\em Proceedings of the IEEE International Conference on Computer
  Vision}, pages 2794--2802, 2015.

\bibitem{weinzaepfel2013deepflow}
P.~Weinzaepfel, J.~Revaud, Z.~Harchaoui, and C.~Schmid.
\newblock Deepflow: Large displacement optical flow with deep matching.
\newblock In {\em Proceedings of the IEEE International Conference on Computer
  Vision}, pages 1385--1392, 2013.

\bibitem{wiskott2002slow}
L.~Wiskott and T.~J. Sejnowski.
\newblock Slow feature analysis: Unsupervised learning of invariances.
\newblock {\em Neural computation}, 14(4):715--770, 2002.

\bibitem{yosinski2014transferable}
J.~Yosinski, J.~Clune, Y.~Bengio, and H.~Lipson.
\newblock How transferable are features in deep neural networks?
\newblock In {\em Advances in Neural Information Processing Systems}, pages
  3320--3328, 2014.

\bibitem{zhang2016color}
R.~Zhang, P.~Isola, and A.~Efros.
\newblock Colorful image colorization.
\newblock In {\em ECCV}, 2016.

\bibitem{placesdata}
B.~Zhou, A.~Lapedriza, J.~Xiao, A.~Torralba, and A.~Oliva.
\newblock Learning deep features for scene recognition using places database.
\newblock {\em NIPS}, 2014.

\end{thebibliography}
% SUPPLEMENTARY
\normalsize

\end{document}